%% file: acl2023.tex
\pdfoutput=1

\documentclass[11pt]{article}

\usepackage{{ACL2023}}

\usepackage{times}
\usepackage{latexsym}

\usepackage[T1]{fontenc}

\usepackage[utf8]{inputenc}

\usepackage{microtype}

\usepackage{inconsolata}

\usepackage[utf8]{inputenc} %
\usepackage[T1]{fontenc}    %
\usepackage{hyperref}       %
\usepackage{url}            %
\usepackage{xspace}
\usepackage{booktabs}       %
\usepackage{amsfonts}       %
\usepackage{nicefrac}       %
\usepackage{microtype}      %
\usepackage{xcolor}         %

\usepackage{amsmath}
\usepackage{bm}
\usepackage{mathtools}
\usepackage{amsthm}
\usepackage{algorithm}
\usepackage{algorithmic}
\usepackage{wrapfig}
\usepackage{enumitem}

\usepackage{graphicx} %
\usepackage{subcaption}
\usepackage[font=small]{caption}
\usepackage{multicol}
\usepackage{multirow}
\usepackage{booktabs}
\usepackage{colortbl}
\usepackage[most]{tcolorbox}

\newcommand{\method}{\texttt{CoPlanner}\xspace}

\title{Cooperative Strategic Planning Enhances Reasoning Capabilities in Large Language Models}

\author{
 \textbf{Danqing Wang\textsuperscript{1}\thanks{Correspondence to \texttt{danqingw@cs.cmu.edu}.}},
 \textbf{Zhuorui Ye\textsuperscript{2}},
 \textbf{Fei Fang\textsuperscript{1}},
 \textbf{Lei Li\textsuperscript{1}}
\\
 \textsuperscript{1}Carnegie Mellon University,
 \textsuperscript{2}Tsinghua University,
}

\begin{document}
\maketitle

\begin{abstract}
\input{sec/000abstract}

\end{abstract}

\section{Introduction}
\label{sec:intro}
\input{sec/001introduction}

\section{Related Work}
\label{sec:related}
\input{sec/002related}

\section{Cooperative Reasoning Framework}
\label{sec:method}
\input{sec/003method}

\section{Experiment}
\label{sec:expr}
\input{sec/004experiment}

\section{Conclusion}
\label{sec:conclusion}
\input{sec/005conclusion}

\section*{Limitation}
\input{sec/007limitation}

\section*{Ethics Statement}
\input{sec/006ethic}

\bibliography{anthology,custom}
\bibliographystyle{acl_natbib}

\newpage

\appendix

\section{Appendix}
\label{sec:appendix}
\input{sec/010appendix}

\end{document}

%% file: sec/000abstract.tex
Enhancing the reasoning capabilities of large language models (LLMs) is crucial for enabling them to tackle complex, multi-step problems. Multi-agent frameworks have shown great potential in enhancing LLMs' reasoning capabilities. However, the lack of effective cooperation between LLM agents hinders their performance, especially for multi-step reasoning tasks. This paper proposes a novel cooperative multi-agent reasoning framework (\method) by separating reasoning steps and assigning distinct duties to different agents. \method consists of two LLM agents: a planning agent and a reasoning agent. The planning agent provides high-level strategic hints, while the reasoning agent follows these hints and infers answers. By training the planning agent's policy through the interactive reasoning process via Proximal Policy Optimization (PPO), the LLaMA-3-8B-based \method outperforms the previous best method by 9.94\% on LogiQA and 3.09\% on BBH. Our results demonstrate that the guidance from the planning agent and the effective cooperation between the agents contribute to the superior performance of \method in tackling multi-step reasoning problems.

%% file: sec/001introduction.tex
Recently, research has shown that multiple LLM-based agents can enhance the capability of a single LLM through communication, especially in solving reasoning problems~\citep{du2023improving,hao-etal-2023-reasoning,zhu-etal-2023-solving}. However, when faced with complex reasoning tasks involving multiple reasoning steps, it remains challenging for these agents to find the correct answer. 

The main reason lies in the fact that these agents attempt to directly solve the complex problem independently and communicate with others after obtaining a solution, such as in LLM debate~\citep{du2023improving} or ChatEval~\citep{chan2024chateval}. This setting still poses the challenge of problem-solving to a single agent and even increases the difficulty by requiring the single agent to evaluate others' intricate solutions. Recent studies have found that these multi-agent frameworks do not exhibit significant advantages over elaborated instructions or demonstrations~\citep{huang2023large,wang2024rethinking}. To leverage the benefits of the multi-agent system, we should allow each agent to focus on their respective strengths and attempt to solve only a part of the problem at a time. Based on this approach, they can collaborate by leveraging their individual capabilities and collectively solve the entire problem in a step-by-step manner.

\begin{figure*}
    \centering
    \includegraphics[width=1\linewidth]{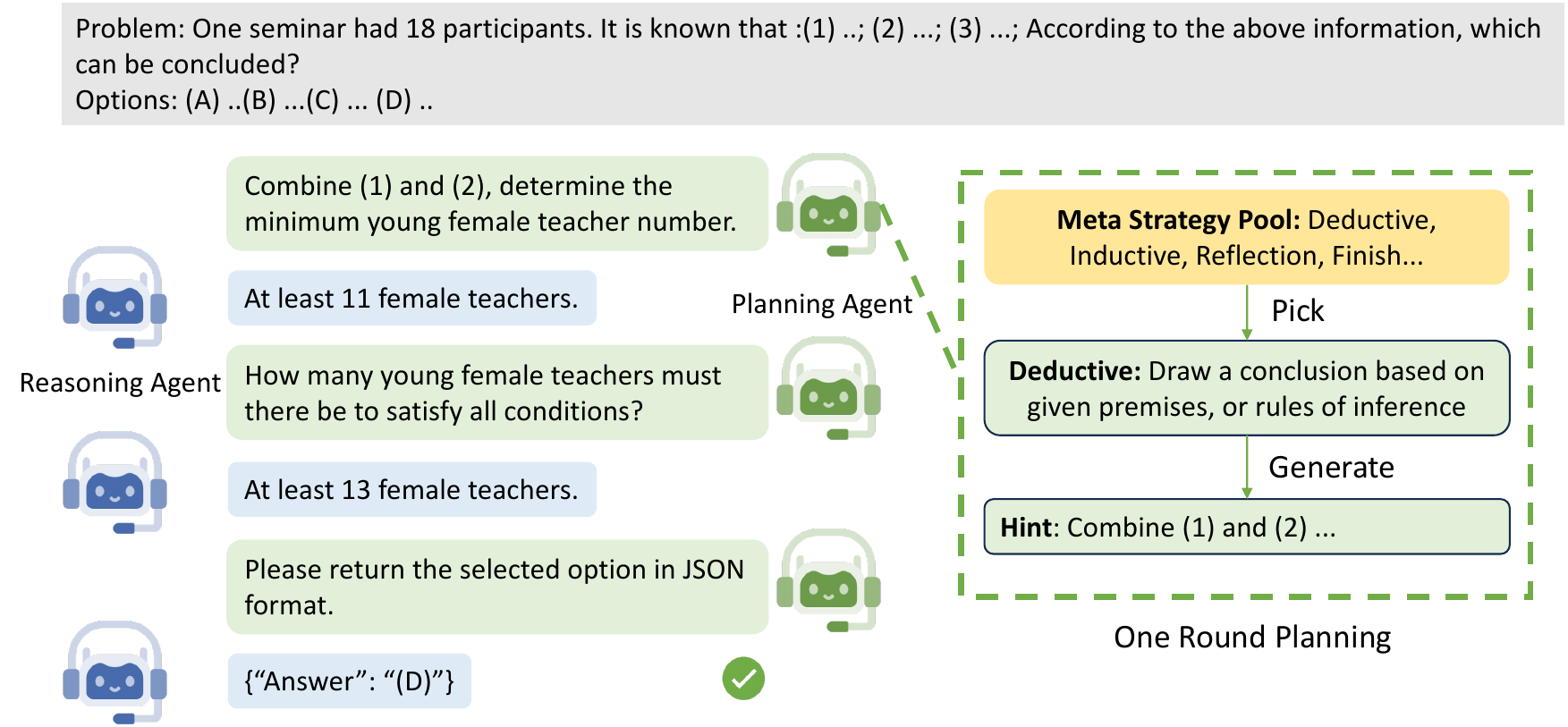}
    \caption{\method consists of two key agents: the reasoning agent and the planning agent. The reasoning agent is responsible for conducting the reasoning process, while the planning agent provides strategic guidance. For each round, the planning agent selects the most appropriate meta-strategy from a pool of meta-strategies based on the historical reasoning process of the reasoning agent. It then generates a detailed hint based on the chosen meta-strategy. This hint is passed to the reasoning agent to guide the next step of reasoning. This interactive process between the two agents continues until the planning agent selects the "finish" strategy, indicating that the reasoning process is complete, or until a maximum number of rounds is reached.}
    \label{fig:overview}
\end{figure*}

In this paper, we introduce a cooperative multi-agent reasoning framework (\method) to enhance reasoning capabilities. \method consists of two agents: a planning agent that provides high-level ideas for problem-solving, and a reasoning agent that focuses on following the plan to perform reasoning. Instead of attempting to solve the problem in a single step, these two agents interact with each other over several rounds, with each round focusing on a specific part of the overall problem. This separation of responsibilities and reasoning steps allows the reasoning agent to concentrate on correctly interpreting and following the immediate one-step instruction provided by the planning agent, without needing to handle the higher-level planning and decomposition of the overall problem.

As illustrated in Figure \ref{fig:overview}, for each round, the planning agent generates a concrete hint for one-step reasoning by selecting a generic meta-strategy from a pre-defined pool and extending it to a problem-specific hint. The meta-strategy pool includes several common problem-solving methodologies such as deduction, induction, reflection, etc. The reasoning agent takes this hint and implements the step-wise instruction. The planning agent then considers the implementation results and provides the next hint.

To facilitate the planning agent's high-level planning capabilities, we employ Proximal Policy Optimization (PPO)~\citep{schulman2017proximal} to learn the policy based on the interaction between the two agents. Specifically, the planning agent comprises a policy network to select the best meta-strategy, a value network to estimate the future reward of the current state, and an LLM to generate the concrete hint. We continuously update the value network and policy network during training and directly apply the learned policy network on the test set. Our key contributions can be summarized as follows:

\begin{itemize}[noitemsep, leftmargin=1em, topsep=1pt]
    \item We propose a novel cooperative multi-agent framework, \method, for complex reasoning tasks. It is composed of a reasoning agent and a planning agent. By separating complex tasks into several steps and assigning responsibilities to different agents, \method takes better advantage of multi-agent cooperation.
    \item We employ behavioral cloning initialization and difficulty-aware curriculum filtering to improve training efficiency, stability, and agent performance by focusing on informative problems that provide meaningful feedback.
    \item LLaMA-3-8B-based and Mistral-7B-based \method outperform other baselines on the LogiQA benchmark, while the Mistral-7B-based variant achieves comparable results to the tree-of-thought policy on BBH with significantly lower inference time.
\end{itemize}

%% file: sec/002related.tex
\textbf{Reasoning in Language Models}
Large Language Models (LLMs) have demonstrated impressive logical reasoning and problem-solving capabilities. Approaches like Chain-of-Thought (CoT) \citep{wang2022self} enhance reasoning by prompting LLMs to generate step-by-step solutions, while Tree of Thought (ToT) \citep{yao2023tree} and Graph of Thought \citep{besta2024graph} provide more structured reasoning processes. Aggregating multiple LLM responses can further improve performance \citep{wang2022self}. Recently, researchers have explored using Monte Carlo Tree Search (MCTS) to find the most promising reasoning paths \citep{liu2023making,wang2024promptagent,hao-etal-2023-reasoning}.

\noindent \textbf{Multi-agent in LLMs}
To enhance the capability of solving complicated problems, a branch of work has investigated multi-agent cooperation among LLMs. One approach trains a verifier or reward model to evaluate the reasoning steps generated by a separate LLM \citep{zhu-etal-2023-solving,lightman2023let}. Other studies incorporate an auxiliary LLM to provide natural language feedback, helping the main LLM reflect on and correct mistakes \citep{wang2023learn,akyurek2023rl4f,wang2024tpd}. Another line of work introduces multiple LLMs that debate with each other to improve reasoning \citep{du2023improving,liang2023encouraging,yin2023exchange} or obtain better evaluations \citep{chern2024can,chan2024chateval}. Multi-agent frameworks have also been explored in domains such as games \citep{xu2023language,xu2023exploring}, software development \citep{qian2023communicative}, and real-world simulations \citep{hua2023war}. In this paper, we propose a novel multi-agent reasoning framework that disentangles the responsibilities of planning and implementation for complex tasks. By assigning dedicated agents for high-level planning and focused reasoning, our framework aims to leverage the complementary strengths of different agents and facilitate effective cooperation through structured interactions.

%% file: sec/003method.tex
We begin by formally defining the multi-step reasoning task within the context of our multi-agent framework and provide an overview of our proposed \method. We then delve into the specific roles and responsibilities of the two key agents in \method, namely the reasoning agent and the planning agent. Finally, we describe the training methodology employed to facilitate effective cooperation between these agents and enhance their collective reasoning capabilities.

\subsection{Overview}
We introduce a multi-agent framework for solving complex reasoning tasks step by step. For a given query $\bm{x}$, this framework lets several LLM-based agents discuss several rounds and thus get a solution for this task. 

Specifically, we introduce two agents in \method: \textbf{Planning agent} provides high-level hints $\{\bm{a}_0, \cdots, \bm{a}_t\}$ while \textbf{Reasoning Agent} follows the plan to propose the step-wise solution $\{\bm{y}_0, \cdots, \bm{y}_t\}$. $t \in [0,\cdots, T]$ indicates the $t-$th round and $\bm{y}_t$ is the final answer.

We define the planning agent as a Markov decision process~(MDP) $(\mathcal{S}, \mathcal{A}, \mathcal{P}, \mathcal{R})$. The state $\bm{s}_t \in \mathcal{S}$ includes the given query $\bm{x}$ and the historical thoughts of the reasoning agents $\{y_0, \cdots, y_t\}$. The action $\bm{a}_t \in \mathcal{A}$ is a hint from several LLM-generated hints based on a pre-defined meta-strategy pool $\mathcal{C}$. $\mathcal{P}$ describes the transition $(\bm{s}_t, \bm{a}_t) \to \bm{s}_{t+1}$, which depends on the generation of the reasoning agent after giving the new hint $\bm{a}_t$. The reward is calculated based on the correctness of the final answer $y_t$. If it is correct, the reward is 1 otherwise 0.

At each round $t$, the planning agent selects a meta-strategy $c_t$ based on the current problem-solving state $\bm{s}_t$ and generates a concrete hint $\bm{a}_t$ based on the query and historical thoughts of the reasoning agent. The reasoning agent follows the hint $h_t$ to conduct one-step reasoning $y_t$. When the planning agent chooses to stop the reasoning process, or when the maximum round $T$ is reached, the reasoning agent is forced to give a final answer.

Through the collaboration between the planning agent and the reasoning agent, each with distinct responsibilities, we decompose complex problem-solving into a multi-step process. This disentangles the implementation of the planning. The reasoning agent's sole responsibility is to accurately follow the immediate instruction provided by the planning agent, while the planning agent handles the higher-level planning and decomposition of the overall problem into a sequence of steps.

\subsection{Planning Agent for Strategic Planning}
The goal of the planning agent is to choose the best hint for the reasoning agent to help it get the correct answer. It includes three modules: a value network $v_{\theta}(\bm{s}_t)$ to approximate the reward of the current state, a policy network $p_{\theta}(c_t|\bm{s}_t)$ to select the best meta-strategy to explore, and a language model $\text{LM}(\bm{s}_t,c)$ to provide the concrete hint $\bm{a}_t$. For each round, the planning agent will

\begin{enumerate}[noitemsep, topsep=2pt]
    \item Receive the thoughts $\bm{y}_{t-1}$ from the reasoning agent and update the current state $s_t$.
    \item Select the best meta-strategy $c_t$ from the pre-defined meta-strategy pool $\mathcal{C}$ based on the current state: $p_{\theta}(c_t|\bm{s}_t)$.
    \item Generate the concrete hint based on the meta-strategy: $\bm{a}_t = \text{LM}(\bm{s}_t,c_t)$ and send the hint to the reasoning agent.
    \item If the maximum round is reached, terminate the interaction by asking the reasoning agent to return the answer.
\end{enumerate}

\noindent \textbf{Meta-strategy Pool}
We pre-define a set of 10 meta-strategies based on human cognition. It includes five common logical reasoning types~\cite{wason1972psychology}: \textit{deduction, induction, abduction, analogy, contradiction}, and four problem-solving methods: \textit{decomposition, enumeration, elimination, and reflection}. We also add a meta-strategy \textit{finish} to indicate the end of the reasoning. The detailed instructions are listed in Appendix \ref{sec:a_meta_strategy}. These high-level strategies make the hints more diverse and thus enhance the exploration of the solutions. 

\noindent \textbf{Concrete Hint Generation}
The selected meta-strategy is extended to a concrete hint that is related to the specific query $\bm{x}$ and previous thoughts $\{\bm{y}_0, \cdots, \bm{y}_{t-1}\}$. For example, the hint driven by the meta-strategy reflection can further point out the mistakes in previous thoughts and ask the reasoning agent to revise. The concrete hint makes the instruction easier to follow.

\subsection{Reasoning Agent for Problem-solving}
The goal of the reasoning agent is to provide a step-wise reasoning path $\{\bm{y}^0, \cdots, \bm{y}_T\}$ for the given query. It follows the hint from the planning agent and conducts detailed reasoning to get results. It relies on an instruction-tuned language model to generate language response $\bm{y}_t = \text{LM}(\bm{s}_t, \bm{a}_t)$. Specifically, the reasoning agent will

\begin{enumerate}[noitemsep, topsep=2pt]
    \item Receive the hints $\bm{a}_{t-1}$ from the planning agent.
    \item Conduct one-step reasoning based on the given hint: $\bm{y}_t = \text{LM}(\bm{s}_t, \bm{a}_t)$.
    \item Output the final answer based on previous thoughts once receive the finish signal.
\end{enumerate}

\begin{figure}
    \centering
    \includegraphics[width=1\linewidth]{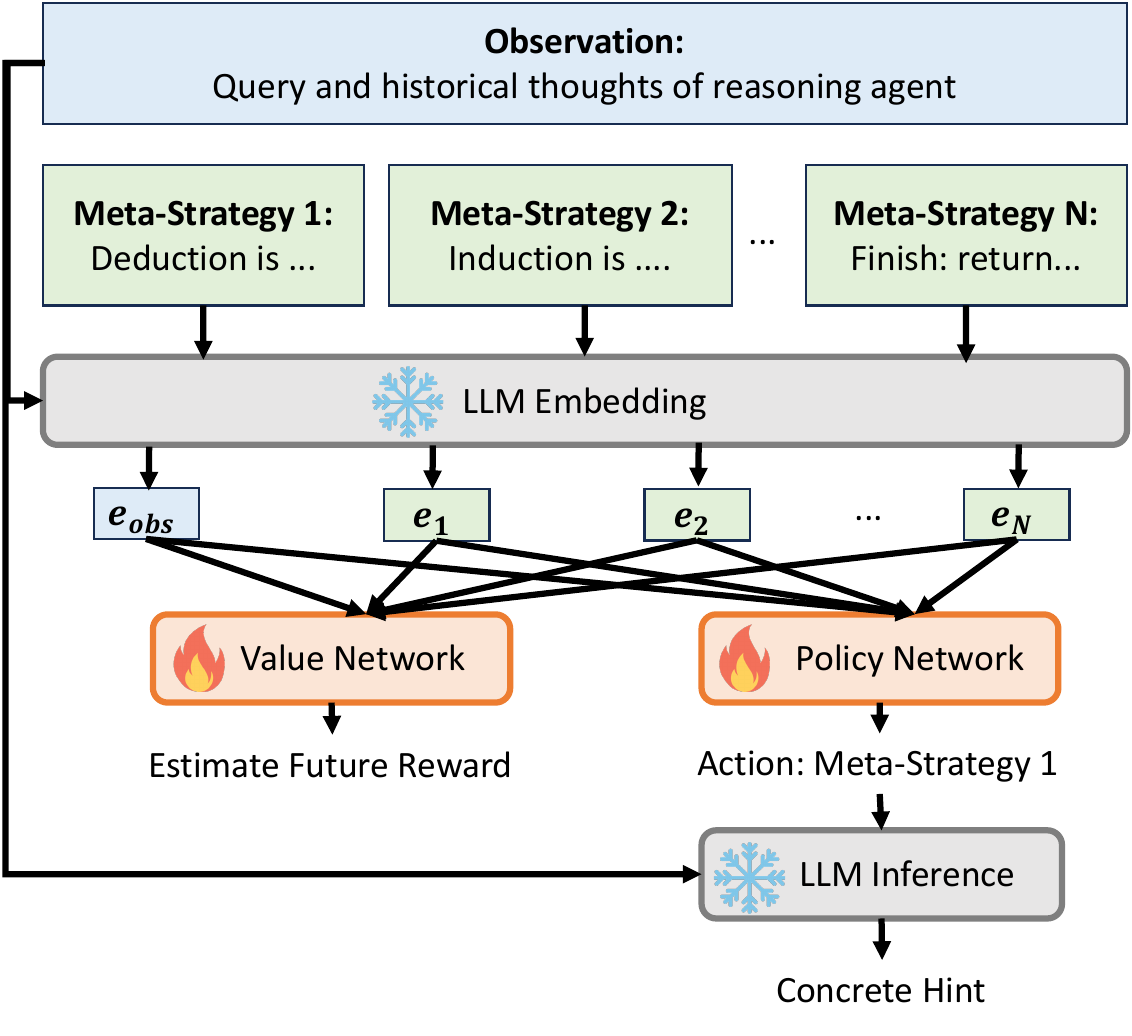}
    \caption{The detailed diagram of the planning agent. It takes the query and historical thoughts of the reasoning agent as the observation and obtains several candidate strategies based on the meta-strategy pool. The observation and candidate strategies use the last hidden state of a frozen LLM as their representation. We use PPO to train a critic network to approximate the reward and a policy network to select the best meta-strategy.}
    \label{fig:module}
\end{figure}

\subsection{Update Planning Policy via Interaction}
We use reinforcement learning to help the planning agent learn the policy of selecting the best hints from the candidate pools. Specifically, we use Proximal Policy Optimization (PPO)~\citep{schulman2017proximal} to train the policy network. It is an actor-critic method that introduces a value network $v_{\theta}(\bm{s}_t)$ to estimate the expected future rewards and updates the policy network $p_{\theta}(c|\bm{s}_t)$ to maximize towards the estimated reward. 

As demonstrated in Figure \ref{fig:module}, we use a frozen LLM to get the embedding of the observation $\bm{s}_t$ and the candidate actions in the meta-strategy pool. The embedding is based on the mean pooling over the tokens' hidden states from the last layer. We use the concatenated embedding as the input of the value and policy network. The value network is composed of one transformer layer~\citep{NIPS2017_3f5ee243} and a linear layer for reward prediction. The policy network uses one transformer layer and outputs the attention weight between the observation and the action as the actions' probabilities. We sample from the action probability distribution during the training and use the best action during the inference. After obtaining the meta-strategy, we use the frozen LLM to generate a concrete hint. 

During training, we set up the interaction between the reasoning and the planning agent as described above. After the reasoning agent provides its final answer, we compute the reward $R_t$ by comparing the answer with the ground truth. We assign reward = 1 if the answer is correct and -1 otherwise. We use the observed reward $R_t$ and the current state $s_t$ to update the value network $v_{\theta}(s_t)$ via temporal-difference learning. We compute the advantage estimate $A_t$ using the value network's predictions and the observed rewards and update the policy network $p_{\theta}(c_t|\bm{s}_t)$ by the PPO objective function: 
\begin{align}
    J^{\text{PPO}}(\theta) = \mathbb{E}_{t}[
    & \min(r_t(\theta) A_t, \nonumber \\
    & \text{clip}(r_t(\theta), 1-\epsilon, 1+\epsilon)A_t)], \\
    & r_t(\theta) = \frac{p_{\theta}(c_t|s_t)}{p_{\theta_{\text{old}}}(c_t|s_t)}.
\end{align}
Here $\epsilon$ is a hyperparameter that enforces the new policy $p_\theta$ remains close to the old policy $p_{\theta_{\text{old}}}$

\noindent \textbf{Initialization via Behavior Cloning}
We collect successful trajectories to initialize the policy network via behavior cloning. Specifically, for each round, the policy agent randomly selects one meta-strategy and generates the hints for the reasoning agent, while the reasoning agent follows the given hints to infer the answer. We collect the interaction between two agents, and sample 32 times for each example in the training set. We evaluate the answer at the end of the interaction and only the trajectories with correct answers are kept. These trajectories reformulated as state-action pairs $D_{bc} = \{(\bm{s}_t, \bm{a}_t) | t \in \{0, \cdots, T\}\}$. We use cross entropy to maximize the likelihood of the action $\bm{a}_t$ selected in state $\bm{s}_t$. This policy is named as $p_{\theta_0}(c_t|s_t)$.

\noindent \textbf{Difficulty-Aware Curriculum Filtering}
Inspired by curriculum learning, we curate the set of problems used for training based on their difficulty to improve the training efficiency and effectiveness. We observe that simple problems contribute less to the RL training because they get a reward of 1 most of the time. Meanwhile, the most challenging ones always receive negative rewards, hindering the training. Therefore, we sort the training problems by their difficulty and filter the easiest and hardest ones. We estimate the difficulty of each problem by the successful trajectories collected in the behavior cloning. The difficulty of example $i$ is defined as $\delta_i = \frac{\text{\# Success(i)}}{\text{\# Sample(i)}}$. $\text{\# Success(i)}$ is the number of successful trajectories of $i$ and  $\text{\# Sample(i)}$ is the sample number, which is 32 for all examples. This selective approach enables the reasoning agent to learn from informative problems and helps to stabilize the training process. By focusing on problems that provide meaningful feedback and opportunities for improvement, we can optimize the interaction between the reasoning agent and the planning agent, leading to more effective policy updates and enhanced overall performance.

In this online PPO setting, the planning agent continuously updates its value network and policy network during its interactions with the reasoning agent. The updates are based on the observed rewards and states, allowing the planning agent to adapt and improve its hints over time.

%% file: sec/004experiment.tex
\subsection{Datasets}
We use two multi-choice reasoning benchmarks that require multiple reasoning steps. The reasoning agent is asked to return the answer in JSON format. We calculate the accuracy based on the exact match between the predicted answer and the ground truth. 
LogiQA~\citep{10.5555/3491440.3491941,10174688} is a multi-choice understanding benchmark for logical reasoning. We follow the standard training/validation split and only keep examples with more than 3 reasoning categories. This makes the problem more diverse and difficult to solve. This results in 1517 training examples and 203 test examples. We take the validation set as the test set and randomly select 200 examples from the training set for validation.
BBH~\citep{suzgun2022challenging} is a set of hard problems borrowed from Big Bench~\citep{srivastava2022beyond}. They are also formatted as multi-choice problems. We pick the English tasks with more than 2 options, resulting in 16 tasks\footnote{They are date understanding, disambiguation qa, geometric shapes, hyperbaton, logical deduction three, logical deduction five, logical deduction seven, movie recommendation, penguins in a table, reasoning color, ruin names, snarks, temporal sequences, tracking shuffled three, tracking shuffled five, and tracking shuffled seven.}. For each task, we randomly select 200 examples as the test set and 20 examples as the validation. The rest are used as training examples. 

\subsection{Implementation Details}
We use Mistral 7B Instruct-v0.2~\citep{jiang2023mistral} and LLaMA 3 8B Instruct ~\cite{touvron2023llama1,jiang2023mistral} as our backbone models. We use a fine-tuned mistral embedding model~\citep{wang2023improving} to get the representation. The hidden size of the transformer layer is 64 and one attention head is used in the policy and value network. 
We collect 88,441 and 17,200 state-action pairs for LogiQA and BBH. The data are randomly split by 9:1 for training and validation. We use the learning rate $1e-4$ and 16 batch size for behavior cloning and train the police network for 10k steps. The 10-category classification accuracy on the validation set is 0.776 on LogiQA and 0.605 on BBH. We keep problems with $\delta_i \in [5\%, 90\%]$ difficulty filtering.
For PPO training, we set the $\epsilon=0.1$. The batch size is 32, and the training epoch of PPO is 10. After initializing with the behavior policy $p_{\theta_0}$ we first freeze the policy network for 1k training steps to train the value network. We then use the same learning rate $5e-4$ for the policy and value network with a linear learning decay. The total environment step is 5k. We use one A6000 GPU for training and two A6000 GPUs for LLM inference. The maximum round is set to 2 in the main experiments. More details can be found in Appendix \ref{sec:a_training}.

\subsection{Baselines}
We add several baselines such as the \textbf{Direct}, \textbf{Few-shot}, chain-of-thought (\textbf{CoT}) prompting~\citep{weichain}. For multi-agent settings, we use different approaches as the planning agent. \textbf{Random Policy} randomly picks a meta-strategy from the meta-strategy pool and generates the concrete hint from it. This can be viewed as an ablation of \method with a randomly initialized policy network. \textbf{CoT Policy} prompts the LLM in the planning agent to select the best meta-strategy. We also elaborate a competitive baseline \textbf{ToT Policy} based on \citet{yao2023tree}. We use the tree search to explore the best hint for each round via several prompt-based reward functions. At each round, it samples 3 random hints and self-evaluates these hints. We define three aspects to evaluate the hint (rationality, relevancy, and clarity) and two aspects to evaluate the reasoning steps (correctness and consistency). The hint with the highest average score over five aspects will be selected for the next round~\footnote{The detailed implementation can be found in Appendix \ref{sec:a_prompt}}.

\input{tab/main}

\subsection{Main Results}
We present our main results in Table \ref{tab:main}. The LLaMA-3-8B-based \method outperforms all other baselines on both the LogiQA and BBH benchmarks. The Mistral-7B-based \method demonstrates comparable performance to the Tree of Thought (ToT) policy on BBH but achieves better results on LogiQA. However, the ToT policy requires significantly more time during inference compared to \method. It evaluates each candidate hint based on the hint itself and the reasoning result based on that hint, resulting in a runtime three times longer than \method, even with batch generation. Furthermore, the quality of the selected hint highly relies on self-evaluation, making its performance unstable across different LLMs. These results provide empirical evidence of the effectiveness of our \method, while also highlighting several key findings.

\noindent \textbf{Chain-of-Thought Prompting Alone Cannot Help LLMs Pick the Right Meta-Strategy.}
From the comparison between the random policy and the Chain-of-Thought (CoT) policy, we observe that randomly picking a meta-strategy sometimes performs better than using CoT prompting. This indicates that LLMs do not gain the capability of high-level planning during fine-tuning or reinforcement learning with human feedback alone. They excel at following concrete instructions but struggle with strategic thinking. However, when facilitated with reinforcement learning based on the interaction with the reasoning agent, the planning agent learns a policy to select suitable meta-strategies for reasoning based on the reasoning agent's responses.

\noindent \textbf{Multiple Agents Do Not Always Benefit Reasoning.}
We also observe that multi-agent reasoning does not necessarily lead to better performance, especially when the backbone LLM is powerful. Low-quality feedback or plans can mislead the LLM and cause it to reverse correct answers to incorrect ones. This finding is consistent with recent studies suggesting that most interactions in multi-agent frameworks only perturb the prompt instead of providing new ideas, making them easily surpassed by well-designed instructions or demonstrations \citep{huang2023large,wang2024rethinking}. This highlights the importance of high-quality communication in multi-agent frameworks. Our \method disentangles the responsibilities of the two agents, allowing each agent to focus on its strengths and provide high-quality communication between agents.

\subsection{Analysis}
To further investigate \method's performance, we conduct more experiments from several aspects. Unless otherwise stated, experiments are based on Mistral 7B with 2 rounds and 5k training steps, initialized with behavior cloning.

\input{tab/ablation}

\noindent \textbf{Ablation Study}
We remove the key components in \method to investigate their influence. \textbf{w/o Filter} removes the filter of difficulty and trains \method with all data. \textbf{w/o PPO} removes the PPO training and use the behavior cloning policy $p_{\theta_0}$. \textbf{w/o BC} removes the initialization and train PPO from scratch. We list the performance degradation percentage in Table \ref{tab:ablation}. We can see that BC policy has the greatest impact on LogiQA while on BBH its benefit is limited. This may be because the BC policy on LogiQA is much higher than BBH (0.776 v.s. 0.605), providing the PPO with a better initialization. Additionally, difficulty-aware filtering contributes a lot to PPO training by stabilizing the reward signal and making the training process more efficient. Without the filtering, the PPO will perform worse than the BC policy.

\begin{figure}
    \centering
    \includegraphics[width=0.9\linewidth]{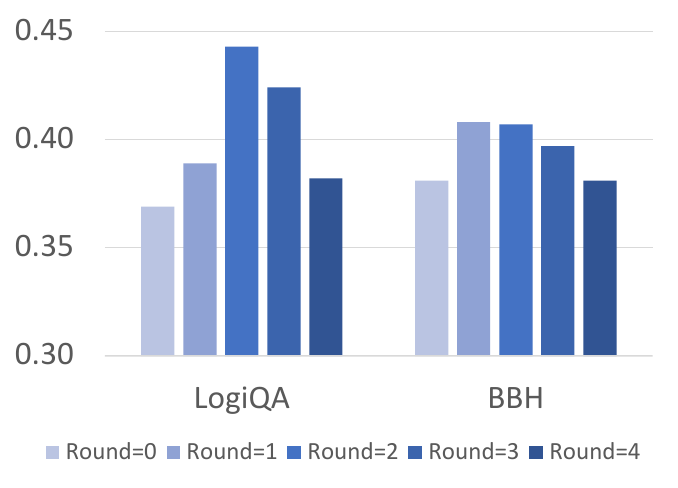}
    \caption{Performance with different interactive rounds between agents. 2 rounds achieve the best performance.}
    \label{fig:round}
\end{figure}

\noindent \textbf{Influence of Interactive Round Number}
We investigate the performance of different rounds in Figure \ref{fig:round}. When round=0, we directly prompt the reasoning agent to solve the problem. For round=$k$, we let the two agents interact at most $k$ round. If the reasoning agent has not reached the final answer, we force the planning agent to provide a \textit{finish} hint. We can see that when the round number is small, the reasoning agent struggles with complex problems. However, when the round number is large, it also confuses the reasoning agent with redundant interaction. Besides, it is more difficult to learn the policy with more rounds, which can also lead to an unsatisfactory performance.

\input{tab/mode}

\noindent \textbf{Variations of the Guide Behavior}
In the main experiments, we let the planning agent pick the meta-strategy and generate a concrete hint based on it. The planning agent can also first generate several hints for each meta-strategy and then select the best hint from them. This will lead to a dynamic action space $\mathcal{A}_t = \{\text{LM}(\bm{s}_t, c) | c \in \mathcal{C}\}$ at each round. We define the policy network on the dynamic meta-strategy pools as $p_{\varphi}(\bm{a}_t | \bm{s}_t)$. We name the planning agent with this dynamic action space as the \textbf{Pick Hint}, and the main experiment setting as \textbf{Pick Meta-strategy}. For these two modes, we can further design two ablation studies. \textbf{Pick Hint w/o Meta-strategy} randomly generate 10 hints with temperature=1, without specifying a meta-strategy. \textbf{Pick Meta-strategy w/o Hint} feeds the generic meta-strategy to the reasoning agent without extending it to a concrete hint. From Table \ref{tab:mode} we can find that the \textbf{Pick Meta-strategy w/o Hint} has the lowest accuracy, indicating that the generic meta-strategy does not have enough details to guide the reasoning agent. Additionally, the randomly generated hints can even perform better than the ones guided by different meta-strategies. This may be because current LLMs do not have enough capability to follow a specific meta-strategy for planning, limiting their capability of thinking from diverse aspects like humans.

\begin{figure}
    \centering
    \includegraphics[width=0.9\linewidth]{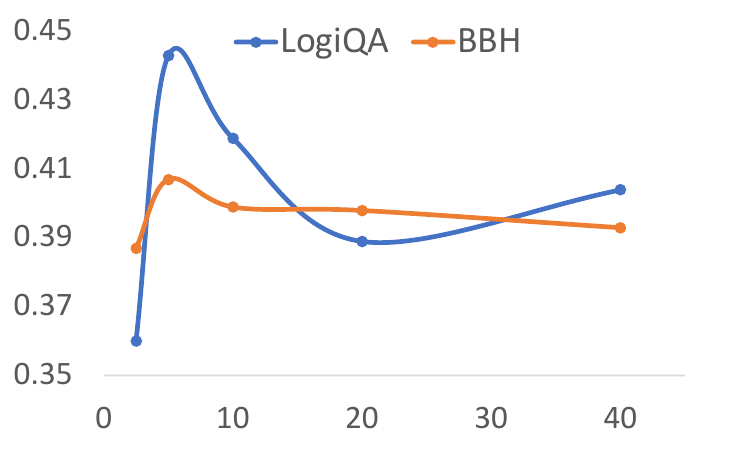}
    \caption{Performance after training with different training steps. The unit of x-axis is thousand.}
    \label{fig:timestep}
\end{figure}

\input{tab/case}

\noindent \textbf{Timestep for PPO Training}
To investigate how many timesteps are enough for PPO, we train PPO with 2.5k, 5k, 10k, 20k, and 40k steps and plot the smoothed figure in Figure \ref{fig:timestep}. The PPO learns very quickly at the beginning, but the performance degrades after 5k steps. However, we can see that when the training step is larger than 20k, the performance of LogiQA starts to increase. We believe it is because the initialization of BC policy makes it easy for PPO training to learn some easy policy patterns on top of it. When it starts to explore more and update the policy, it diverges too much from the initial policy. However, when the timestep is scaled up enough, it may be able to learn more complex policy patterns and achieve better performance.

\textbf{Case Study}
We demonstrate one typical reasoning problem that requires multiple steps in Table \ref{tab:case}. Although the chain-of-thought prompt enforces the LLM to think step by step, the LLM still skips several reasoning steps and infers the answer on a false assumption (\textit{since A lied about ...}). On the other hand, we can see that CoT policy without learning has an obvious preference for \textit{Enumeration}. On the Logiqa test set, it selects the enumerate strategy almost half the time. It may be because of the nature of the multi-choice problem, where we can always check each option. However, for this problem, the inefficient enumeration increases the complexity, leading to a worse result.
However, the learned policy in the planning agent of \method can select a more efficient meta-strategy (\textit{Contradiction}) for this problem. It further asks the reasoning agent to reflect on its previous conclusion in the second round to ensure correctness. The suitable meta-strategy and concrete hints make it easier for the reasoning agent to conduct reasoning, avoiding mistakes caused by skipping steps and untimely reflection.

%% file: tab/main.tex
\begin{table}[htbp]\footnotesize\setlength{\tabcolsep}{4pt}
  \centering
  \caption{\method outperforms both single- and multi-agent baselines with LLaMA 3 8B model. For Mistral 7B, \method achieves comparable results with the competitive ToT policy on BBH but costs less time.}
    \begin{tabular}{lcccc}
    \toprule
          & \multicolumn{2}{c}{Mistral 7B} & \multicolumn{2}{c}{LLaMA 3 8B} \\
          \midrule
          & LogiQA & BBH & LogiQA & BBH \\
    \midrule
    Direct Prompt & 0.369 & 0.381 & 0.493 & 0.440 \\
    Fewshot Prompt & 0.414 & 0.389 & 0.438 & 0.474 \\
    CoT Prompt & 0.399 & 0.375 & 0.458 & 0.475 \\
    \midrule
    Random Policy & 0.394 & 0.388 & 0.429 & 0.486 \\
    CoT Policy & 0.369 & 0.369 & 0.453 & 0.419 \\
    ToT Policy & 0.424 & \textbf{0.409} & 0.389 & 0.401 \\
    \method &  \textbf{0.443} & 0.407 & \textbf{0.542} & \textbf{0.501} \\
    \bottomrule
    \end{tabular}%
  \label{tab:main}%
\end{table}%

%% file: tab/ablation.tex
\begin{table}[htbp]\small
  \centering
  \caption{Ablation on \method's components. The difficulty filtering benefits the PPO training a lot.}
    \begin{tabular}{lcc}
    \toprule
     & LogiQA & BBH \\
    \midrule
    \method & 0.443 & 0.407 \\
    \quad w/o Filter & -8.80\% & -1.72\% \\
    \quad w/o PPO & -6.55\% & -1.23\% \\
    \quad w/o BC & -9.55\%	& -0.74\% \\
    \bottomrule
    \end{tabular}%
  \label{tab:ablation}%
\end{table}%

%% file: tab/mode.tex
\begin{table}[htbp]\small
  \centering
  \caption{Performance of different planning modes. While the meta-strategy with concrete hints has the best performance, the selection of randomly generated hints shows promising results.}
    \begin{tabular}{lcc}
    \toprule
     & LogiQA & BBH \\
    \midrule
    Pick Hint & 0.404 & 0.377  \\
    \quad w/o Meta-strategy & 0.414 & 0.394 \\
    \midrule
    Pick Meta-strategy & 0.443 & 0.407 \\
    \quad w/o Hint & 0.394 & 0.354 \\
    \bottomrule
    \end{tabular}%
  \label{tab:mode}%
\end{table}%

%% file: tab/case.tex
\begin{table*}[th]\small
  \centering
  \caption{One typical case from LogiQA. The prompts and responses are abbreviated. We use round=2 for the Mistral-7B-based \method and omit the finish strategy to terminate the interaction. We represent the concrete hint generated by the planning agent and additionally attach the selected meta-strategy (which is italic and underlined).
  The concrete hints provided by \method guide the reasoning and get the correct answer. }
    \begin{tabular}{lp{40em}}
    \toprule
    \textbf{Problem} & A, B, and C sell their products. Among the three products, one is counterfeit, and the other two are genuine. A said: My product is genuine, and B's product is counterfeit; B said: My product is genuine, and C's product is counterfeit; C said: B's product is genuine, and A's product is counterfeit. It is known that one of A, B, and C told all the truth, truth and lie, and the other told all lies, who sells counterfeit goods? \newline{}
    Options: (A) A (B) B (C) C (D) Not sure\\
    \midrule
    \textbf{Target} & (A) \\
    \midrule
    \textbf{CoT Prompt} & B's statement that their product is genuine and C's counterfeit is true, so B does not sell counterfeit goods. However, since A lied about B's product and their own, A must be the one selling the counterfeit goods  \newline{}
    \textbf{Final Answer}: (B) \\
    \midrule
    \textbf{CoT Policy} & 
    \textbf{Planning Agent}: \underline{[\textit{Enumeration}]}: Consider the possibilities of each person telling the truth, half truth, or lies about the other two products. Test each option to determine which one is most likely to be selling the counterfeit product. \newline{}
    \textbf{Reasoning Agent}: If A tells the truth, ... A tells a half-truth: ... Therefore, B is selling a counterfeit product.\newline{}
    \textbf{Planning Agent}: \underline{[\textit{Enumeration}]}: Consider the possibility that one person is lying, and use the given statements to deduce which product is counterfeit based on the contradictions between their claims. \newline{}
    \textbf{Reasoning Agent}: Consider the possibility that one person is lying... This means that B is selling the counterfeit product. \newline{}
    \textbf{Final Answer}: (D) Not sure \\
    \midrule
    \textbf{\method} & 
    \textbf{Planning Agent}: \underline{[\textit{Contradiction}]} Consider the contradictions between their statements and determine which one could be telling the truth based on their claims about the other two products. \newline{}
    \textbf{Reasoning Agent}: If A is telling the truth, then B's statement that B's product is genuine and C's product is counterfeit is a lie ... consistent with A's statement.
    If B is telling the truth,... contradicts A's statement.
    If C is telling the truth, ... So if A is telling the truth, then the statements are consistent. \newline{}
    \textbf{Planning Agent}: \underline{[\textit{Reflection}]} Consider which statement is consistent with all given information if one seller is telling the truth. Analyzing A's statement. \newline{}
    \textbf{Reasoning Agent}: If A is telling the truth, then the statements are consistent. \newline{}
    \textbf{Final Answer}: (A) \\
    \bottomrule
    \end{tabular}%
  \label{tab:case}%
\end{table*}%

%% file: sec/005conclusion.tex
In this paper, we propose a novel cooperative multi-agent framework (\method) to enhance large language models with strategic planning for complex reasoning tasks. \method decomposes the problem into high-level planning and focused reasoning, assigning these responsibilities to distinct agents. The planning agent learns an effective hint selection policy through PPO on its interactions with the reasoning agent during training. The behavior cloning and the difficulty-aware curriculum filtering make the training more stable and efficient.
Extensive experiments on LogiQA and BBH benchmarks demonstrate the effectiveness of \method. By leveraging the cooperation between multiple agents with specialized roles, \method provides a promising approach to enhancing the reasoning capabilities of large language models.

%% file: sec/007limitation.tex
Although \method demonstrates a novel and effective way to enhance LLM reasoning by cooperation, there are still several limitations. First, this paper mainly focuses on reasoning tasks, and \method's performance on other tasks such as math problems or real-world planning is also interesting. Besides, due to the computation limitation, we focus on 7B models and do not scale up the PPO training to a large scale. It would be interesting to see how the scaling laws of training time and model size affect \method's performance.

%% file: sec/006ethic.tex
We acknowledge that there might be some ethical considerations in enhancing LLMs' reasoning capability such as the \method presented in this paper. However, we believe that no one must be specifically highlighted here.

%% file: sec/010appendix.tex
\subsection{Meta-strategy Instruction}
\label{sec:a_meta_strategy}

\input{tab/prompt}

We list the instructions used for each meta-strategy:
\begin{itemize}[noitemsep, topsep=2pt]
    \item \textbf{Decomposition}: Decompose the problem or the preceding step into easier-to-solve parts.
    \item \textbf{Enumeration}: Enumerate all potential candidates in the context of the given conditions and find the most promising one.
    \item \textbf{Elimination}: Eliminate options that are incorrect or have a very low possibility of being correct.
    \item \textbf{Reflection}: Review previous results and verify whether these results are correct. If not, find the error and correct it.
    \item \textbf{Finish}: Please return the selected option in JSON format. 
    \item \textbf{Deductive Reasoning}: Draw a conclusion based on general truths, principles, given premises, or rules of inference.
    \item \textbf{Inductive Reasoning}: Start from a set of individual instances and generalize to arrive at a general conclusion.
    \item \textbf{Abductive Reasoning}: Make an educated guess based on the known information and verify this guess.
    \item \textbf{Analogical Reasoning}: Start from information about one system and infer information about another system based on the similarity between the two systems.
    \item \textbf{Contradiction}: Demonstrate that a statement is false by assuming it's true and then showing this leads to an impossible or absurd outcome. 
\end{itemize}

\subsection{Prompts}
\label{sec:a_prompt}

\textbf{Tree-of-thought Baseline}
We list the prompts for evaluating the hint quality:
\begin{itemize}[noitemsep, topsep=2pt]
    \item \textbf{Rationality}: Evaluate whether the current hint is a reasonable instruction to solve the problem. 1 is unreasonable, 3 is reasonable, and 2 is unsure. Return "The score is x", where x is an integer from 1 to 3.
    \item \textbf{Relevant}: Evaluate whether the current hint is relevant to the input problem. 1 is irrelevant, 3 is relevant, and 2 is unsure. Return "The score is x", where x is an integer from 1 to 3.
    \item \textbf{Clarity}: Evaluate whether the current hint is easy to understand and follow. 1 is difficult to understand and follow, 3 is easy to understand and follow, and 2 is unsure. Return "The score is x", where x is an integer from 1 to 3.
\end{itemize}

And here are the prompts for the reasoning results.
\begin{itemize}[noitemsep, topsep=2pt]
    \item \textbf{Correctness}: Evaluate whether the answer of the current reasoning hint is correct. 1 is incorrect, 3 is correct, and 2 is unsure. Return "The score is x", where x is an integer from 1 to 3.
    \item \textbf{Consistency}: Evaluate whether the current response is consistent with the input query and the given instruction hint. 1 is inconsistent, 3 is consistent, and 2 is unsure. Return "The score is x", where x is an integer from 1 to 3.
\end{itemize}

\input{tab/hyperparameter}

\subsection{Training Details}
\label{sec:a_training}
\begin{figure}[htbp]
    \centering
    \includegraphics[width=0.85\linewidth]{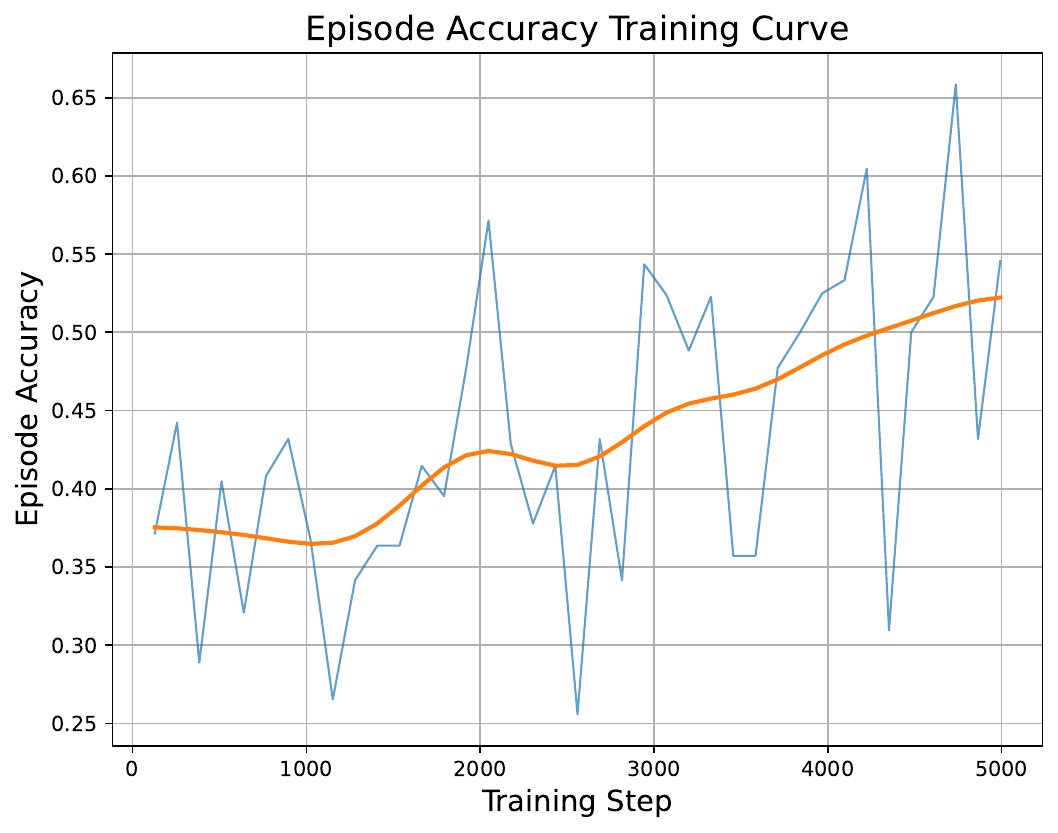}
    \caption{\method's episode accuracy on LogiQA. The blue curve is the raw data while the orange is the smoothed plot. }
    \label{fig:logiqa_train}
\end{figure}

\begin{figure}[htbp]
    \centering
    \includegraphics[width=0.85\linewidth]{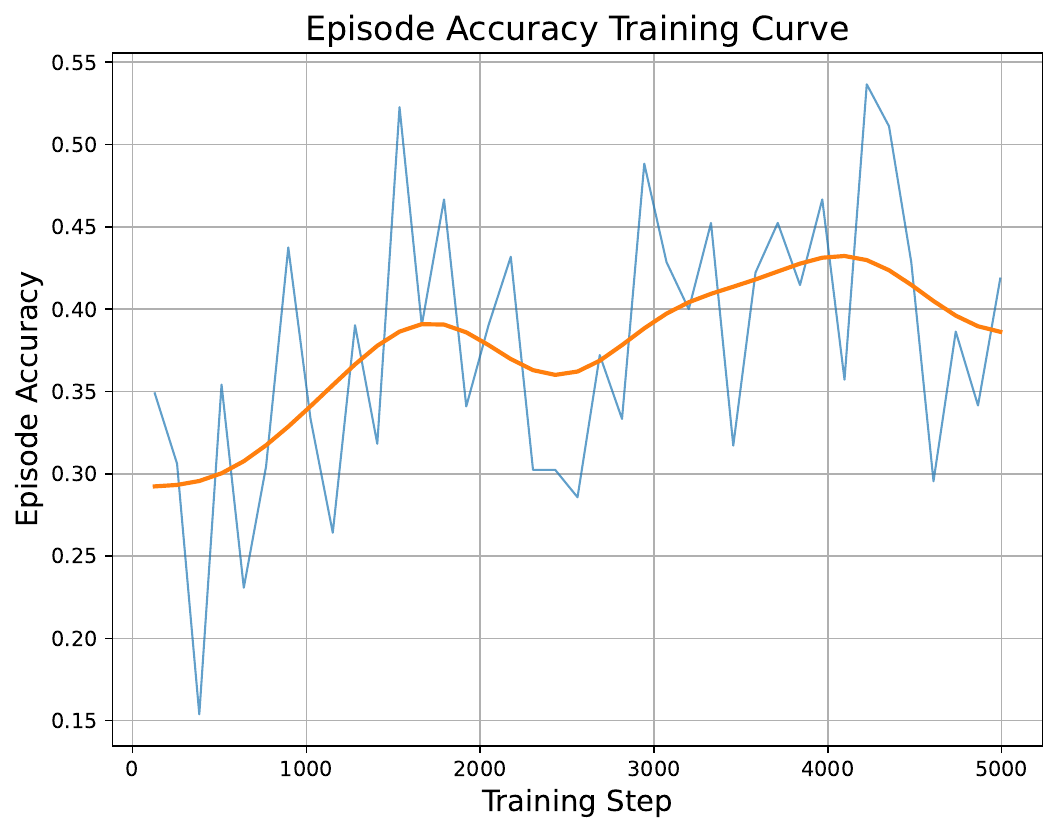}
    \caption{\method's Episode accuracy on BBH.}
    \label{fig:bbh_train}
\end{figure}

In our PPO implementation, the hyperparameters for RL training are listed in Table~\ref{tab:training}. Training speed is around 4 hours for $5000$ timesteps on one A6000 GPU. We randomly pick 100 data on LogiQA and use all data on BBH after the difficulty filtering for PPO training. Note that all training data is used during the behavior cloning. We plot the training curve in Figure \ref{fig:logiqa_train} and \ref{fig:bbh_train}. We use the accuracy of the training data as the y-axis.

%% file: tab/prompt.tex
\begin{table*}[htbp]\small
  \centering
  \caption{Prompts used for concrete hint generation of the planning agent, and the reasoning step of the reasoning agent.}
    \begin{tabular}{lp{35em}}
    \toprule
    \textbf{Hint Generation} & 
    Problem: [query]\newline{}
    Thoughts: [thoughts]\newline{}
    Refer to the given meta-strategy: [strategy]\newline{}\newline{}
    Prepare one potential succeeding hint for the input based on the above strategy. The hint should be brief and begin with 'Hint: '. Do not include the thought process or the result within the hint. For example, the hint for Enumeration can be "Hint: enumerate the options to find the correct answer. Let's start with Option (A)".  \\
    \midrule
    \textbf{One-step Reasoning} & 
    Problem: [query]\newline{}
    Thoughts: [thoughts]\newline{}
    Hint: [suggestion]\newline{}\newline{}
    Let's follow a systematic approach by considering the hint. The previous thoughts are outlined above for reference. \\
    \bottomrule
    \end{tabular}%
  \label{tab:prompt}%
\end{table*}%

%% file: tab/hyperparameter.tex
\begin{table}[htbp]\small
    \centering
    \caption{Hyper-parameters in RL training of \method. }
    \begin{tabular}{cc}
    \toprule
    \textbf{Hyper-parameters }                        & \textbf{Value} \\
    \midrule
    Learning rate                            & 5e-4  \\
    Learning rate decay                      & True  \\
    Discount rate ($\gamma$)                 & 0.99  \\
    GAE parameter ($\lambda_{\textrm{GAE}}$) & 0.95  \\
    Gradient clipping                        & 10.0  \\
    Value loss coefficient                   & 0.5     \\
    Entropy coefficient                      & 1e-5  \\
    PPO clipping                             & 0.1   \\
    PPO epochs                               & 10    \\
    Attention layer head num                 & 1    \\
    Action candidate num $N$                 & 10     \\
    temperature $T$ of the reasoner                 & 0     \\
    \bottomrule
    \end{tabular}
    \label{tab:training}
\end{table}